\def\year{2020}\relax
\documentclass[letterpaper]{article} %
\usepackage{aaai20}  %
\usepackage{times}  %
\usepackage{helvet} %
\usepackage{courier}  %
\usepackage[hyphens]{url}  %
\usepackage{graphicx} %
\usepackage{graphicx}  %
\usepackage[utf8]{inputenc}
\usepackage{amsmath}
\usepackage{amsfonts}
\usepackage{amssymb}
\usepackage{multirow}
\usepackage{color}
\usepackage{bm}
\usepackage{placeins}
\usepackage{pifont}
\usepackage{graphicx}
\usepackage{enumitem}
\usepackage{physics}
\graphicspath{{./figures/}}

\def\eg{\emph{e.g.~}} 
\def\ie{\emph{i.e.~}}

\newcommand{\cmark}{\ding{51}}
\newcommand{\xmark}{\ding{55}}

\title{Distribution-Aware Coordinate Representation for Human Pose Estimation}
\author{Feng Zhang$^1$
	\quad \quad \quad \quad Xiatian Zhu$^2$
	\quad \quad \quad \quad Hanbin Dai$^1$
	\quad \quad \quad \quad Mao Ye$^1$
	\quad \quad \quad \quad Ce Zhu$^1$
	\\
	University of Electronic Science and Technology of China$^1$
	\quad \quad \quad \quad \quad \quad 
	University of Surrey$^2$\\
	{\tt\small \{zhangfengwcy, eddy.zhuxt, daihanbin.ac, cvlab.uestc\}@gmail.com}
	\quad \quad \quad
	\tt\small eczhu@uestc.edu.cn
}

\begin{document}

\maketitle

\begin{abstract}
While being the {\em de facto} standard coordinate representation
in human pose estimation,
{\em heatmap} is never systematically investigated in the literature, to our best knowledge.
This work fills this gap by
studying the coordinate representation with a particular 
focus on the heatmap.
Interestingly, we found that
the process of {\em decoding} the predicted heatmaps into the final joint coordinates
in the original image space is {\em surprisingly significant}
for human pose estimation performance,
which nevertheless was not recognised before.
In light of the discovered importance,
we further probe the design limitations
of the standard coordinate decoding method widely used
by existing methods, and propose a more principled 
distribution-aware decoding method. 
Meanwhile, we improve the standard coordinate {\em encoding} process
(\ie transforming ground-truth coordinates to heatmaps)
by generating accurate heatmap distributions for unbiased model training.
Taking the two together, we formulate a novel 
{\em Distribution-Aware coordinate Representation of Keypoint} (DARK) method.
Serving as a model-agnostic plug-in, DARK significantly improves the performance 
of a variety of state-of-the-art human pose estimation models.
Extensive experiments show that DARK yields the best results on two common benchmarks,
MPII and COCO,
consistently validating the usefulness and effectiveness of our novel coordinate representation idea. 
The project page is at \url{https://ilovepose.github.io/coco/}
\end{abstract}

\section{Introduction}

\begin{figure}[ht]
	\centering
	\includegraphics[scale=0.5]{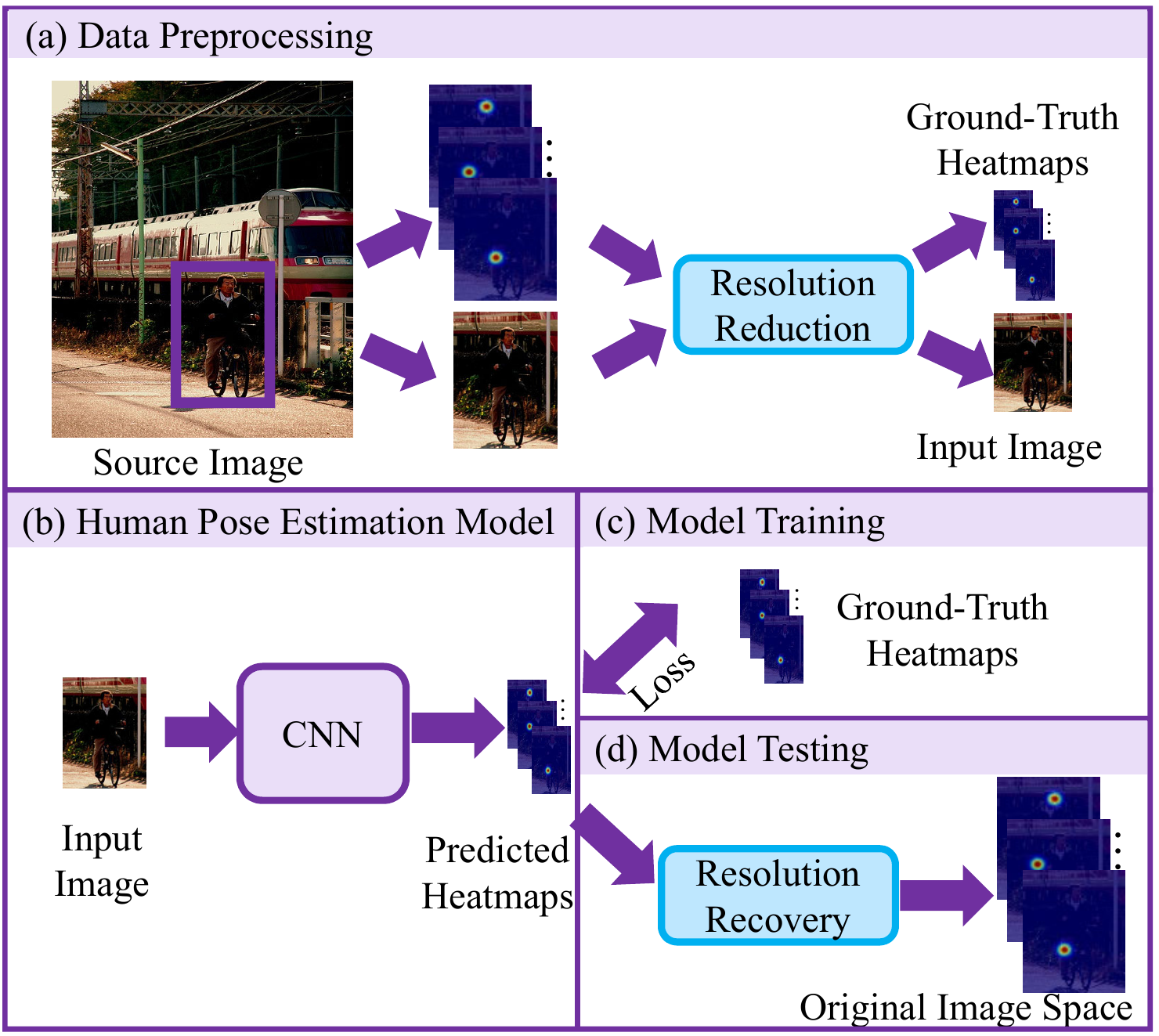}
	\caption{
		Pipeline of a human pose estimation system.
		For efficiency, 
		{\em resolution reduction} is often applied
		on the original person detection bounding boxes
		as well as the ground-truth heatmap supervision.
		That is, the model operates in a low-resolution image space.
		At test time, a corresponding {\em resolution recovery}
		is therefore necessary in order to obtain the joint coordinate prediction in the original image space.
	}
	\label{fig:pipeline}
\end{figure}

Human pose estimation is a fundamental computer vision problem that aims to detect the {\em spatial location} (\ie {\em coordinate}) of human body joints in unconstrained images \cite{andriluka14cvpr}. 
It is a non-trivial task
as the appearance of body joints
vary dramatically due to diverse  
styles of clothes, arbitrary 
occlusion, and
unconstrained background contexts,
whilst it is needed to identify the {\em fine-grained}
joint coordinates.
As strong image processing models,
convolutional neural networks (CNNs)
excel at this task \cite{lecun1998gradient}. %
Existing works typically 
focus on designing the CNN architecture
tailored particularly for human pose inference 
\cite{newell2016stacked,sun2019deep}.

Analogous to the common {\em one-hot vectors} as the {object class label representation}
in image classification, 
a human pose CNN model also requires a \textbf{\em label representation} for encoding the {\em body joint coordinate labels}, so that the supervised learning
loss can be quantified and computed
during training and the joint coordinates can be inferred properly\footnote{
	The {\em label representation} is for encoding the label annotations (\eg 1,000 one-hot vectors for 1,000 object class labels in ImageNet), totally different from the {\em data representation} for encoding the data samples 
	(\eg the object images from ImageNet).}.
The {\em de facto} standard label representation
is {\em coordinate heatmap},
generated as a 2-dimensional Gaussian 
distribution/kernel centred at the labelled coordinate
of each joint
\cite{tompson2014joint}.
It is obtained from a \textbf{\em coordinate encoding} process,
{\em from coordinate to heatmap}.
Heatmap is characterised by giving spatial support around the ground-truth location, considering not only the contextual clues
but also the inherent target position ambiguity.
Importantly, this may effectively reduce
the model overfitting risk in training,
in a similar spirit of
the class label smoothing regularisation
\cite{szegedy2016rethinking}.
Come as no surprise, the state-of-the-art pose models
\cite{newell2016stacked,xiao2018simple,sun2019deep}
are based on the heatmap coordinate representation.

With the heatmap label representation, 
one major obstacle is that, the computational cost is 
a {\em quadratic} function of {\em the input image 
	resolution}, preventing the CNN models
from processing the typically {\em high-resolution} raw imagery data.
To be computationally affordable,
a standard strategy (see Fig. \ref{fig:pipeline}) is to downsample all the person
bounding box images at arbitrarily large resolutions into a prefixed small resolution with a data preprocessing procedure, before being fed into a human pose estimation model.
Aiming to predict the joint location
in the {\em original} image coordinate space,
after the heatmap prediction 
a corresponding {\em resolution recovery} is required
for transforming back to the original coordinate space.
The final prediction is considered as the 
location with the maximal activation.
We call this process as \textbf{\em coordinate decoding},
{\em from heatmap to coordinate}.
It is worthy noting that quantisation error 
can be introduced during the above resolution reduction. %
To alleviate this problem,
during the existing coordinate decoding process a hand-crafted shifting operation is usually performed 
according to  %
the direction from the highest activation to the second highest activation
\cite{newell2016stacked}.

In the literature, the problem of coordinate encoding and decoding (\ie denoted as {\em coordinate representation}) gains little attention,
although being indispensable in model inference. %
In contrast to the current research focus
on designing more effective CNN structures,
we reveal a {\em surprisingly} important role the coordinate representation plays
on the model performance, much more significant than expected.
For instance, with the state-of-the-art model
HRNet-W32 \cite{sun2019deep}, the aforementioned shifting operation of coordinate encoding
brings as high as 
5.7\% AP on the challenging COCO validation set
(Table \ref{tbl:coord_decod}).
It is noteworthy to mention that, this gain is already much more significant than those
by most individual art methods.
But it
is never well noticed and carefully investigated in the literature to our best knowledge.

Contrary to 
the existing human pose estimation studies,
in this work we dedicatedly investigate the problem 
of joint coordinate representation including encoding and decoding.
Moreover, we recognise that the heatmap resolution 
is one major obstacle that prevents the use of smaller 
input resolution for faster model inference.
When decreasing the input resolution
from 256$\times$192 to 
128$\times$96,
the model performance of HRNet-W32 drops significantly
from 74.4\% to 
66.9\% %
on the COCO validation set,
although the model inference cost 
falls
from 7.1$\times 10^{9}$
to 1.8$\times 10^{9}$ FLOPs.

In light of the discovered significance of
coordinate representation,
we conduct in-depth investigation and recognise that
one key limitation lies in the coordinate decoding 
process.
Whilst existing standard shifting operation has shown to be 
effective as found in this study, we propose a principled distribution-aware representation method
for more accurate joint localisation at sub-pixel accuracy. 
Specifically, it is designed to comprehensively account for 
the distribution information of heatmap activation
via Taylor-expansion based distribution approximation.
Besides, we observe that the standard
method for generating the ground-truth heatmaps %
suffers from {\em quantisation errors},
leading to imprecise supervision signals and
inferior model performance.
To solve this issue,
we propose generating the {\em unbiased} heatmaps allowing
Gaussian kernel being centred at sub-pixel locations.

The {\bf contribution} of this work is that,
we discover the previously unrealised significance of coordinate representation
in human pose estimation,
and propose a novel {\em Distribution-Aware coordinate Representation of Keypoint} (DARK) method with two key components:
(1) efficient Taylor-expansion based coordinate decoding,
and 
(2) unbiased sub-pixel centred coordinate encoding.
Importantly, existing human pose methods can be seamlessly
benefited from DARK
{\em without} any algorithmic modification.
Extensive experiments on two common benchmarks (MPII and COCO)
show that our method provides significant performance improvement
for existing state-of-the-art human pose estimation models
\cite{sun2019deep,xiao2018simple,newell2016stacked},
achieving the best single model accuracy 
on COCO and MPII. %
DARK favourably enables the use
of smaller input image resolutions with much smaller 
performance degradation, whilst 
dramatically boosting the model inference efficiency
therefore facilitating low-latency and low-energy applications
as required in embedded AI scenarios.

\section{Related Work}
There are two common coordinate representation
designs in human pose estimation:
direct coordinate and heatmap.
Both are used as the regression targets for model training.

\vspace{0.1cm}
{\noindent \bf Coordinate regression }
Directly taking the coordinates as model output target
is straightforward and intuitive.
But only a handful of existing methods adopt this design
\cite{toshev2014deeppose,fan2015combining,carreira2015human,sun2018integral}.
One plausible reason is that,
this representation 
lacks the spatial and contextual information,
making the learning of human pose model 
extremely challenging due to the intrinsic
visual ambiguity in joint location.

\vspace{0.1cm}
{\noindent \bf Heatmap regression }
The heatmap representation elegantly addresses the above limitations.
It was firstly introduced in \cite{tompson2014joint} %
and rapidly became the most commonly used coordinate representation.
Generally, the mainstream research focus is on designing network architectures
for more effectively regressing the heatmap supervision.
Representative design improvements
include sequential modelling \cite{gkioxari2016chained,belagiannis2016recurrent},
receptive field expansion \cite{wei2016convolutional},
position voting \cite{lifshitz2016human},
intermediate supervision \cite{newell2016stacked,wei2016convolutional},
pairwise relations modelling \cite{chen2014articulated},
tree structure modelling \cite{chu2016crf,yang2016end,chu2016structured,sun2017compositional,tang2018deeply},
pyramid residual learning \cite{yang2017pyramid},
cascaded pyramid learning \cite{chen2018cascaded},
knowledge-guided learning \cite{ning2017tmm},
active learning \cite{liu2017active},
adversarial learning \cite{chen2017adversarial},
deconvolution upsampling \cite{xiao2018simple},
multi-scale supervision \cite{ke2018multi},
attentional mechanism \cite{liu2018cascaded,su2019multi}, 
and high-resolution representation preserving \cite{sun2019deep}.

In contrast to all previous works,
we instead investigate the issues of heatmap representation
on human pose estimation,
a largely ignored perspective in the literature.
Not only do we reveal a big impact
of resolution reduction in the process of using heatmap
but also we propose a principled coordinate
representation method for significantly improving 
the performance of existing models.
Crucially, our method can be seamlessly integrated without model design change.

\section{Methodology}
We consider the coordinate representation problem
including encoding and decoding
in human pose estimation. 
The objective is 
to predict the joint coordinates 
in a given input image.
To that end, we need to learn a regression model
from the input image to the output coordinates,
and the {\em heatmap} is often leveraged as coordinate representation
during both model training and testing.
Specifically, we assume access to a training set 
of images.
To facilitate the model learning, 
we {\em encode} the labelled ground-truth coordinate of a joint %
into a heatmap as the supervised learning target.
During testing, we then need to {\em decode} the predicted heatmap 
into the coordinate in the original image coordinate space.

In the following we first describe the decoding process,
focusing on the limitation analysis of the existing standard method
and the development of a novel solution.
Then, we further discuss and address the limitations of the encoding process.
Lastly, we describe the integration of existing human pose estimation models
with the proposed method.

\subsection{Coordinate Decoding}
Despite being considered as an insignificant component of the model testing pipeline,
as we found in this study, coordinate decoding turns out to be one 
of the most significant performance contributors
for human pose estimation in images
(see Table \ref{tbl:coord_decod}).
Specifically, this is a process of translating
a predicted heatmap of each individual joint
into a coordinate in the {\em original} image space.
Suppose the heatmap has the same spatial size as the original image,
we only need to find the location of the maximal activation as the joint coordinate prediction, which
is straightforward and simple.
However, this is often not the case as interpreted above.
Instead, we need to upsample the heatmaps to the original image resolution
by a sample-specific unconstrained factor $\lambda \in \mathcal{R}_+$.
This involves a {\em sub-pixel localisation} problem.
Before introducing our method, 
we first revisit the standard coordinate decoding method
used in existing pose estimation models.

\vspace{0.1cm}
\noindent{\bf The standard coordinate decoding method}
is designed empirically according to model performance \cite{newell2016stacked}.
Specifically, given a heatmap $\bm{h}$ predicted by a trained model,
we first identify the coordinates of the maximal ($\bm{m}$) and second maximal ($\bm{s}$) activation.
The joint location is then predicted as
\begin{equation}
\bm{p} = \bm{m} + 0.25 \frac{\bm{s}-\bm{m}}{\|\bm{s}-\bm{m}\|_2}
\label{eq:standard_shfit}
\end{equation}
where $\| \cdot \|_2$ defines the magnitude of a vector.
This means that the prediction is as the maximal activation with a 0.25 pixel (\ie sub-pixel) shifting
towards the second maximal activation in the heatmap space.
The final coordinate prediction in the original image is computed as:
\begin{equation}
\hat{\bm{p}} = \lambda \bm{p}
\label{eq:scl}
\end{equation}
where $ \lambda$ is the resolution reduction ratio.

\vspace{0.1cm}
{\em Remarks }
The aim of the sub-pixel shifting in Eq. \eqref{eq:standard_shfit} 
is to compensate the quantisation effect of image resolution downsampling.
That being said, the maximum activation in the predicted heatmap does not 
correspond to the accurate position of the joint in the original coordinate space,
but only to a {\em coarse} location. 
As we will show, this shifting {\em surprisingly} brings a significant performance
boost (Table \ref{tbl:coord_decod}). 
This may partly explain why it is often used as a standard operation in model test.
Interestingly, to our best knowledge no specific work has delved into 
the effect of this operation on human pose estimation performance.
Therefore, its true significance has never been really recognised and reported in the literature.
While this standard method lacks intuition and interpretation in design, no dedicated investigation
has been carried out for improvement.
We fill this gap by presenting a principled method
for shifting estimation and finally more accurate human pose estimation.

\begin{figure*}%
	\centering
	\includegraphics[width=0.95\linewidth]{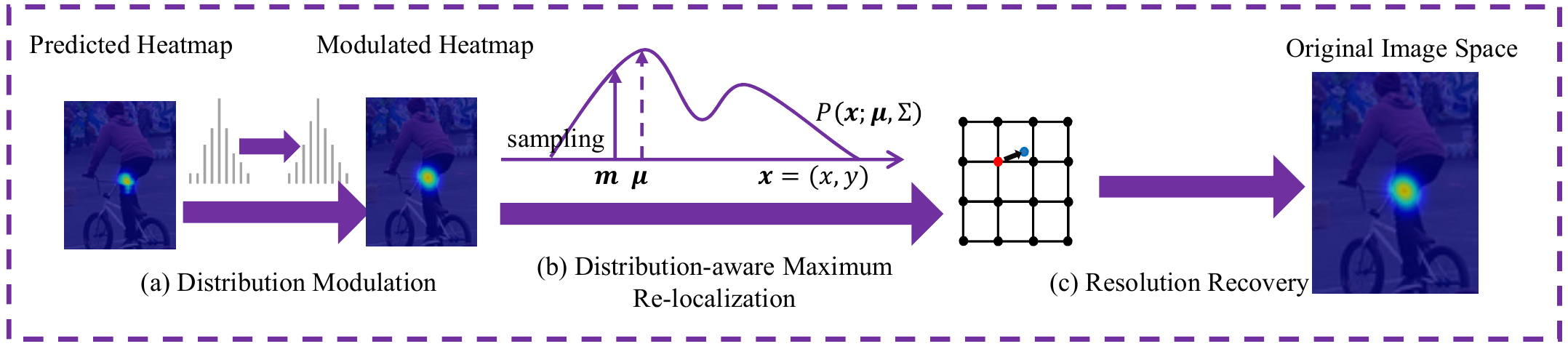}
	\caption{Overview of the proposed distribution aware coordinate decoding method.
	}
	\label{fig:test_stage}
\end{figure*}

\vspace{0.1cm}
\noindent{\bf The proposed coordinate decoding method}
explores the distribution structure of the predicted heatmap to infer the underlying 
maximum activation.
This differs dramatically to the standard method above 
relying on a hand-designed offset prediction,
with little design justification and rationale.

Specifically, to obtain the accurate location at the degree of sub-pixel, 
we assume the predicted heatmap follows a 2D Gaussian distribution,
same as the ground-truth heatmap.
Therefore, we represent the predicted heatmap as
\begin{equation}\small
\mathcal{G} ( \bm{x}; \bm{\mu}, \Sigma ) =
\frac{1}{ (2\pi) |\Sigma|^{\frac{1}{2}} }
\exp( -\frac{1}{2} (\bm{x}-\bm{\mu})^{T} \Sigma^{-1} (\bm{x}-\bm{\mu}) )
\label{eq:gaussian}
\end{equation}
where %
$\bm{x}$ is a pixel location in the predicted heatmap, 
$\bm{\mu}$ is the Gaussian mean (centre) corresponding 
to the {\em to-be-estimated} joint location. %
The covariance $\Sigma$ is a diagonal matrix, same as
that used in coordinate encoding:
\begin{equation}
\Sigma =\begin{bmatrix} 
\sigma^2 & 0 \\
0 & \sigma^2 
\end{bmatrix}
\end{equation}
where $\sigma$ is the standard deviation same for both directions.

In the log-likelihood optimisation principle %
\cite{goodfellow2016deep},
we transform $\mathcal{G}$ through logarithm to facilitate inference while
keeping the original location of the maximum activation
as:
\begin{align}
	\mathcal{P}(\bm{x}; \bm{\mu}, \Sigma) = \ln( \mathcal{G} )%
	= & -\ln(2\pi) - \frac{1}{2}\ln(|\Sigma|) \\ \notag
	& -\frac{1}{2} ( \bm{x} - \bm{\mu} )^{T} \Sigma^{-1} (\bm{x} - \bm{\mu})
	\label{eq:3}
\end{align}

Our objective is to estimate $\bm{\mu}$.
As an extreme point in the distribution, 
it is well-known that the first derivative at the location $\bm{\mu}$ 
meets a condition as: %
\begin{equation}
\eval{\mathcal{D}'(\bm{x})}_{\bm{x} = \bm{\mu}}  = 
\eval{\frac{ \partial{\mathcal{P}^{T} } }{ \partial{\bm{x}} }}_{\bm{x} = \bm{\mu}} 
= \eval{- \Sigma^{-1} (\bm{x} - \bm{\mu})}_{\bm{x} = \bm{\mu}} 
= 0
\label{eq:zero_der}
\end{equation}

To explore this condition, we adopt the Taylor's theorem. 
Formally, we approximate the activation $\mathcal{P}(\mu)$ 
by a Taylor series (up to the quadratic term) evaluated at the maximal activation $\bm{m}$ 
of the predicted heatmap as
\begin{equation}\small
\mathcal{P}(\bm{\mu})  = \mathcal{P}(\bm{m})
+ \mathcal{D}'(\bm{m}) (\bm{\mu} - \bm{m})
+ \frac{1}{2} (\bm{\mu} - \bm{m}) ^T\mathcal{D}''(\bm{m}) (\bm{\mu} - \bm{m})
\label{eq:taylor}
\end{equation}
where $\mathcal{D}''(\bm{m})$ denotes the 
second derivative (\ie Hessian) of $\mathcal{P}$ evaluated at $\bm{m}$,
formally defined as:
\begin{equation}
\mathcal{D}''(\bm{m})
= \eval{\mathcal{D}''(\bm{x})}_{\bm{x} = \bm{m}}  
= - \Sigma^{-1}
\label{eq:hessian}
\end{equation}
The intuition of selecting $\bm{m}$ to approximate $\bm{\mu}$ is that
it represents a good coarse joint prediction that 
approaches $\mathbf{\bm{\mu}}$.

Taking Eq. \eqref{eq:zero_der}, \eqref{eq:taylor}, and \eqref{eq:hessian}
together, we eventually obtain 
\begin{equation}
\bm{\mu}
= \bm{m} - \big(\mathcal{D}''(\bm{m}) \big)^{-1} \mathcal{D}'(\bm{m})
\label{eq:final_joint}
\end{equation}
where $\mathcal{D}''(\bm{m})$ and $\mathcal{D}'(\bm{m})$ can be estimated efficiently 
from the heatmap.
Once obtaining $\bm{\mu}$,
we also apply Eq. \eqref{eq:scl} to predict the
coordinate in the original image space.

\vspace{0.1cm}
{\em Remarks }
In contrast to the standard method considering
the second maximum activation alone in heatmap, 
the proposed coordinate decoding fully explores the heatmap 
distributional statistics for revealing the underlying maximum more accurately.
In theory, our method is based on a principled distribution approximation
under a training-supervision-consistent assumption that the heatmap 
is in a Gaussian distribution.
Crucially, it is very efficient computationally
as it only needs to compute the first and second derivative of one location per heatmap.
Consequently, existing human pose estimation approaches
can be readily benefited without any computational cost barriers.

\begin{figure}
	\centering
	\includegraphics[width=0.99\linewidth]{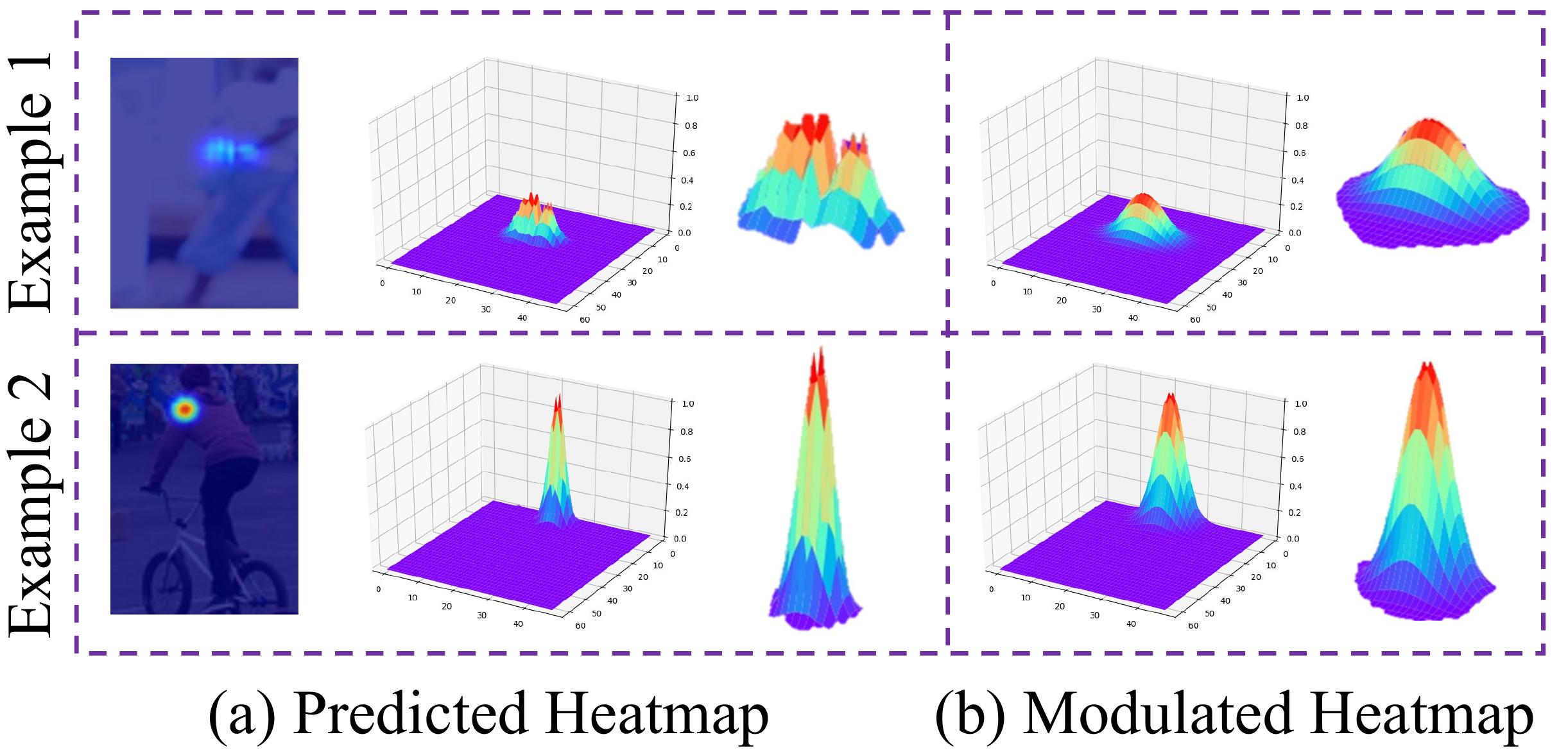}
	\caption{
		Illustration of heatmap distribution modulation.
		(a) Predicted heatmap;
		(b) Modulated heatmap distribution. 
	}
	\label{fig:distribuion}
\end{figure}

\vspace{0.1cm}
\noindent \textbf{\em Heatmap distribution modulation}
As the proposed coordinate decoding method
is based on a Gaussian distribution assumption,
it is necessary for us to examine how well 
this condition is satisfied.
We found that, often, the heatmaps
predicted by a human pose estimation model
do {\em not} exhibit good-shaped Gaussian structure
compared to the training heatmap data. %
As shown in Fig. \ref{fig:distribuion}(a), the heatmap
usually presents multiple peaks around the maximum activation.
This may cause negative effects
to the performance of our decoding method.
To address this issue,
we propose {\em modulating} the heatmap distribution
beforehand. %

Specifically, to match the requirement of our method we 
propose exploiting a Gaussian kernel $K$ with the same variation
as the training data to smooth out the effects of multiple peaks
in the heatmap $\bm{h}$, formally as
\begin{equation}
\bm{h}' = K \circledast \bm{h}
\label{eq:dist_mod}
\end{equation}
where $\circledast$ specifies the convolution operation.

To preserve the original heatmap's magnitude, we finally scale 
$\bm{h}'$ so that its maximum activation is equal to that of $\bm{h}$,
via the following transformation:
\begin{equation}
\bm{h}' = \frac{ \bm{h}' - \min(\bm{h}') }{ \max(\bm{h}') - \min(\bm{h}') } * \max({\bm{h}})
\label{eq:dist_mod_norm}
\end{equation}
where $\max()$ and $\min()$ return the maximum and minimum values
of an input matrix, respectively.
In our experimental analysis,
it is validated that this distribution modulation
further improves the performance of 
our coordinate decoding method (Table \ref{tbl:coord_encode}),
with the resulting visual effect and qualitative evaluation demonstrated in Fig. \ref{fig:distribuion}(b).

\vspace{0.1cm}
\noindent \textbf{\em Summary}
We summarise our coordinate decoding method
in Fig. \ref{fig:test_stage}.
Specifically, a total of three steps are involved in a sequence:
{\bf (a)} Heatmap distribution modulation (Eq. \eqref{eq:dist_mod}, \eqref{eq:dist_mod_norm}),
{\bf (b)} Distribution-aware joint localisation by Taylor expansion
at sub-pixel accuracy (Eq. \eqref{eq:gaussian}-\eqref{eq:final_joint}),
{\bf (c)} Resolution recovery to the original coordinate space (Eq. \eqref{eq:scl}).
None of these steps incur high computational costs,
therefore being able to serve as an efficient plug-in
for existing models.

\begin{figure}%
	\centering
	\includegraphics[width=0.3\linewidth]{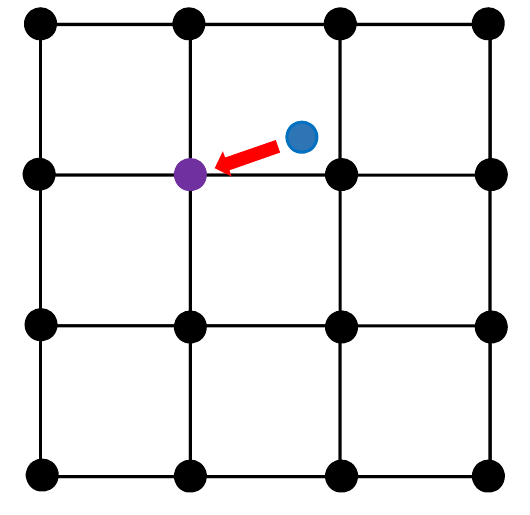}
	\caption{
		Illustration of quantisation error in the standard coordinate encoding process.
		The blue point denotes the accurate position ($\bm{g}'$) of a joint.
		With the {\em floor} based coordinate quantisation, an error (indicated by red arrow) is introduced.
		Other quantisation methods share the same problem.
	}
	\label{fig:quantise}
\end{figure}

\subsection{Coordinate Encoding}
\label{sec:method_heatmap}

The previous section has addressed the problem
with coordinate decoding,
rooted at resolution reduction.
As a similar process, coordinate encoding
shares the same limitation.
Specifically, the standard coordinate encoding method
starts with downsampling original person images
into the model input size. So, the ground-truth joint coordinates
require to be transformed accordingly before generating the heatmaps.

Formally, we denote by $\bm{g}=(u,v)$ the ground-truth coordinate of a joint.
The resolution reduction is defined as:
\begin{align}
	\bm{g}' =(u',v') = \frac{\bm{g}}{\lambda} = (\frac{{u}}{\lambda}, \frac{{v}}{\lambda})
\end{align}
where $\lambda$ is the downsampling ratio.

Conventionally, for facilitating the kernel generation,
we often quantise $\bm{g}'$:
\begin{align}
	\bm{g}'' =(u'',v'') = \text{quantise}(\bm{g}') = \text{quantise}(\frac{{u}}{\lambda}, \frac{{v}}{\lambda})
	\label{eq:quantise}
\end{align}
where $\text{quantise}()$ specifies a quantisation function, with the common choices including
floor, ceil and round.

Subsequently, the heatmap centred at the quantised coordinate $\bm{g}''$ can be
synthesised through:
\begin{equation}\small
\mathcal{G} (x, y;\bm{g}'' ) = 
\frac{1}{ 2\pi \sigma^2 }
\exp( -\frac{(x-u'')^{2} + (y-v'')^2}{2\sigma^2}  )
\label{eq:heatmap_gt}
\end{equation}
where $(x,y)$ specifies a pixel location in the heatmap,
and $\sigma$ denotes a fixed spatial variance.

Obviously, the heatmaps generated in the above way
are {\em inaccurate} and {\em biased} due to the quantisation error (Fig. \ref{fig:quantise}). 
This may introduce sub-optimal supervision signals
and result in degraded model performance,
particularly for the case of accurate coordinate encoding
as proposed in this work.

To address this issue, we simply place the heatmap centre
at the non-quantised location $\bm{g}'$ which represents
the {\em accurate} ground-truth coordinate.
We still apply Eq. \eqref{eq:heatmap_gt} 
but replacing $\bm{g}''$ with $\bm{g}'$.
We will demonstrate the benefits of this {\em unbiased}
heatmap generation method (Table \ref{tbl:coord_encode}).

\subsection{Integration with State-of-the-Art Models}
Our DARK method
is model-agnostic, seamlessly integrable 
with any existing heatmap based pose models.
Importantly, this does not involve any algorithmic changes to 
previous methods.
In particular, during training the only change
is the ground-truth heatmap data generated based 
on the accurate joint coordinates.
At test time, we take as input the predicted heatmaps predicted by
any model such as HRNet \cite{sun2019deep},
and output more accurate joint coordinates in the original image space.
In the whole lifecycle, we keep an existing model intact as the original design.
This allows to maximise the generality and scalability of our method.

\section{Experiments}

{\bf \noindent Datasets}
We used two popular human pose estimation datasets, COCO and MPII.
The \textbf{\em COCO} keypoint dataset \cite{lin2014microsoft} presents
naturally challenging imagery data with various human poses, 
unconstrained environments, different body scales and occlusion patterns.
The entire objective involves both detecting person instances and localising the body joints.
It contains 200,000 images and 250,000 person samples.
Each person instance is labelled with 17 joints. 
The annotations of training and validation sets are publicly benchmarked.
In evaluation, we followed the commonly used train2017/val2017/test-dev2017 split.
The \textbf{\em MPII} human pose dataset \cite{andriluka14cvpr} 
contains 40k person samples, each labelled with 16 joints.
We followed the standard train/val/test split as in \cite{tompson2014joint}.

\vspace{0.1cm}
{\bf \noindent Evaluation metrics }
We used Object Keypoint Similarity (OKS) for COCO and 
Percentage of Correct Keypoints (PCK)
for MPII to evaluate the model performance.

\vspace{0.1cm}
{\bf \noindent Implementation details } 
For model training, we used the Adam optimiser.
For HRNet \cite{sun2019deep} and SimpleBaseline \cite{xiao2018simple}, we followed the same learning schedule and epochs as in the original works.
For Hourglass \cite{newell2016stacked}, the base learning rate was fine-tuned to 2.5e-4,
and decayed to 2.5e-5 and 2.5e-6 at the 90-th and 120-th epoch.
The total number of epochs is 140.
We used three different input sizes ($128\times96$, $256\times192 $, $384\times288 $) in our experiments.
We adopted the same data preprocessing as in \cite{sun2019deep}.

\begin{table}%
	\setlength{\tabcolsep}{0.1cm}
	\begin{center}
		\caption{
			Effect of coordinate decoding on the COCO validation set.
			Model: HRNet-W32;
			Input size: $128\times96$.
		}
		\label{tbl:coord_decod}
		\begin{tabular}{ c | c |c | c | c | c | c  }
			\hline
			Decoding & $AP$ & $AP^{50}$ & $AP^{75}$ & $AP^{M}$ & $AP^{L}$ & $AR$ \\
			\hline \hline
			No Shifting
			& 61.2 & 88.1 & 72.3 & 59.0 & 66.3 & 68.7 
			\\
			Standard Shifting
			& 66.9 &\bf 88.7 & 76.3 & 64.6 & 72.3 & 73.7 
			\\ 
			\hline
			\bf Ours
			& \bf 68.4 & 88.6 & \bf 77.4 & \bf 66.0 & \bf 74.0 & \bf 74.9
			\\
			\hline
		\end{tabular}
	\end{center}

\end{table}

\begin{table}[ht]
	\setlength{\tabcolsep}{0.2cm}
	\begin{center}
		\caption{
			Effect of distribution modulation (DM) on the COCO val set. 
			Backbone: HRNet-W32;
			Input size: 128$\times$96.
		}
		\label{tbl:ablation_GB}
		\begin{tabular}{ c | c |c | c | c | c | c  }
			\hline
			DM & $AP$ & $AP^{50}$ & $AP^{75}$ & $AP^{M}$ & $AP^{L}$ & $AR$ \\
			\hline \hline
			\xmark
			& 68.1 & 88.5 & 77.1 & 65.8 & 73.7 & 74.8
			\\ 
			\hline
			\cmark
			& \bf 68.4 & \bf 88.6 & \bf 77.4 & \bf 66.0 & \bf 74.0 & \bf 74.9
			\\ 
			\hline
		\end{tabular}
	\end{center}
\end{table}

\begin{table} %
	\setlength{\tabcolsep}{0.1cm}
	\begin{center}
		\caption{
			Effect of coordinate encoding on the COCO validation set.
			Model: HRNet-W32;
			Input size: $128\times96$.
		}
		\label{tbl:coord_encode}
		\resizebox{\columnwidth}{!}{%
			\begin{tabular}{ c| c | c |c | c | c | c | c  }
				\hline
				Encode & Decode & $AP$ & $AP^{50}$ & $AP^{75}$ & $AP^{M}$ & $AP^{L}$ & $AR$ \\
				\hline \hline
				Biased & Standard
				& 66.9 & 88.7 & 76.3 & 64.6 & 72.3 & 73.7 
				\\
				\bf Unbiased & Standard
				& \bf 68.0 & \bf 88.9  & \bf 77.0  & \bf 65.4  & \bf 73.7  & \bf 74.5
				\\
				\hline \hline
				Biased &\bf Ours
				& 68.4 & 88.6 & 77.4 & 66.0 & 74.0 & 74.9
				\\
				\bf Unbiased &\bf Ours
				& \bf 70.7 & \bf  88.9 & \bf  78.4 & \bf  67.9 & \bf  76.6 & \bf  76.7
				\\
				\hline
			\end{tabular}
		}
	\end{center}
\end{table}

\begin{table} %
	\setlength{\tabcolsep}{0.1cm}
	\begin{center}
		\caption{
			Effect of input image size on the COCO validation set.
			DARK uses HRNet-W32 (HRN32) as backbone.
		}
		\label{tbl:ablation_input_size}
		\resizebox{\columnwidth}{!}{%
			\begin{tabular}{ c | c | c | c | c | c | c | c | c  }
				\hline
				Method & Input size & GFLOPs & $AP$ & $AP^{50}$ & $AP^{75}$ & $AP^{M}$ & $AP^{L}$ & $AR$ \\
				\hline \hline
				HRN32 & \multirow{2}{*}{128$\times$96} & \multirow{2}{*}{1.8}
				& 66.9 & 88.7 & 76.3 & 64.6 & 72.3 & 73.7 
				\\
				\bf DARK &  &
				& \bf 70.7 & \bf 88.9 & \bf 78.4 & \bf 67.9 & \bf 76.6 & \bf 76.7
				\\ \hline
				HRN32 & \multirow{2}{*}{256$\times$192} & \multirow{2}{*}{7.1}
				& 74.4	& \bf 90.5	& 81.9	& 70.8	& 81.0	& 79.8
				\\ 
				\bf DARK & &
				& \bf 75.6 & \bf 90.5 & \bf 82.1 & \bf 71.8 & \bf 82.8 & \bf 80.8
				\\ \hline
				HRN32 & \multirow{2}{*}{384$\times$288} & \multirow{2}{*}{16.0}
				& 75.8	& 90.6 & 82.5 & 72.0 & 82.7 & 80.9
				\\ 
				\bf DARK & & 
				& \bf 76.6 & \bf 90.7 & \bf 82.8 & \bf 72.7 & \bf 83.9 & \bf 81.5
				\\ 
				\hline
			\end{tabular}
		}
	\end{center}
\end{table}

\subsection{Evaluating Coordinate Representation}
As the core problem in this work,
the effect of coordinate representation on model 
performance was firstly examined, with a connection to the input image resolution (size).
In this test, by default we used %
HRNet-W32 \cite{sun2019deep} as the backbone model and $128\times96$ as the input size,
and reported the accuracy results on the COCO validation set.

\begin{table*} %
	\setlength{\tabcolsep}{0.22cm}
	\begin{center}
		\caption{
			Evaluating the generality of our DARK method to varying state-of-the-art
			models on the COCO validation set. 
		}
		\label{tbl:ablation_generalisation}
		\resizebox{\linewidth}{!}{%
			\begin{tabular}{ c | l | c | c | c | c | c | c | c | c | c }
				\hline
				\bf DARK & Baseline & Input size & \#Params  & GFLOPs & $AP$ & $AP^{50}$ & $AP^{75}$ & $AP^{M}$ & $AP^{L}$ & $AR$ \\
				\hline \hline
				\xmark & \multirow{2}{*}{Hourglass (4 Blocks)} & \multirow{2}{*}{$128\times96$} & \multirow{2}{*}{13.0M} & \multirow{2}{*}{2.7}
				& 66.2 & 87.6 & 75.1 & 63.8 & 71.4 & 72.8
				\\
				\cmark & & & &  
				&\bf 69.6 &\bf 87.8 &\bf 77.0 &\bf 67.0 &\bf 75.4 & \bf 75.7
				\\ \hline
				\xmark & \multirow{2}{*}{Hourglass (8 Blocks)} & \multirow{2}{*}{$128\times96$} & \multirow{2}{*}{25.1M} & \multirow{2}{*}{4.9}
				& 67.6 &\bf 88.3 & 77.4 & 65.2 & 73.0 & 74.0
				\\
				\cmark & & & &  
				&\bf 70.8 & 87.9 &\bf 78.3 &\bf 68.3 &\bf 76.4 &\bf 76.6
				\\ \hline \hline
				\xmark & \multirow{2}{*}{SimpleBaseline-R50} & \multirow{2}{*}{$128\times96$} & \multirow{2}{*}{34.0M} & \multirow{2}{*}{2.3}
				& 59.3 & 85.5 & 67.4 & 57.8 & 63.8 & 66.6
				\\
				\cmark & & & &  
				&\bf 62.6 &\bf 86.1 &\bf 70.4 &\bf 60.4 &\bf 67.9 &\bf 69.5
				\\ \hline
				\xmark & \multirow{2}{*}{SimpleBaseline-R101} & \multirow{2}{*}{$128\times96$} & \multirow{2}{*}{53.0M} & \multirow{2}{*}{3.1}
				& 58.8 & 85.3 & 66.1 & 57.3 & 63.4 & 66.1
				\\
				\cmark & & & &  
				&\bf 63.2 &\bf 86.2 & \bf 71.1 &\bf 61.2 &\bf 68.5 &\bf 70.0
				\\ \hline
				\xmark & \multirow{2}{*}{SimpleBaseline-R152} & \multirow{2}{*}{$128\times96$} & \multirow{2}{*}{68.6M} & \multirow{2}{*}{3.9}
				& 60.7 & 86.0 & 69.6 & 59.0 & 65.4 & 68.0
				\\
				\cmark & & & &  
				&\bf 63.1 &\bf 86.2 &\bf 71.6 &\bf 61.3 &\bf 68.1 &\bf 70.0
				\\ \hline \hline 
				\xmark & \multirow{2}{*}{HRNet-W32} & \multirow{2}{*}{$128\times96$} & \multirow{2}{*}{28.5M} & \multirow{2}{*}{1.8}
				& 66.9 & 88.7 & 76.3 & 64.6 & 72.3 & 73.7
				\\
				\cmark & & & & 
				&\bf 70.7 &\bf 88.9 &\bf 78.4 &\bf 67.9 &\bf 76.6 &\bf 76.7
				\\ \hline
				\xmark & \multirow{2}{*}{HRNet-W48} & \multirow{2}{*}{$128\times96$} & \multirow{2}{*}{63.6M} & \multirow{2}{*}{3.6}
				& 68.0 & 88.9 & 77.4 & 65.7 & 73.7 & 74.7
				\\
				\cmark & & & & 
				&\bf 71.9 &\bf 89.1 &\bf 79.6 &\bf 69.2 &\bf 78.0 &\bf 77.9
				\\ 
				\hline
			\end{tabular}
		}
	\end{center}
\end{table*}

\vspace{0.1cm}
\noindent{\bf (i) Coordinate decoding }
We evaluated the effect of coordinate decoding, in particular, the shifting operation and distribution modulation.
The conventional biased heatmaps were used.
In this test, we compared the proposed distribution-aware shifting method with 
{\em no shifting} (\ie directly using the maximal activation location),
and the {\em standard shifting} (Eq. \eqref{eq:standard_shfit}).
We make two major observations in Table \ref{tbl:coord_decod}:
{\bf (i)} The standard shifting
gives as high as 5.7\% AP accuracy boost,
which is surprisingly effective.
To our best knowledge, this is the first reported effectiveness 
analysis in the literature, since this problem is largely
ignored by previous studies.
This reveals previously unseen significance of coordinate
decoding to human pose estimation.
{\bf (ii)} Despite the great gain by the standard decoding method,
the proposed model further improves AP score by 1.5\%,
among which the distribution modulation gives 0.3\%
as shown in Table \ref{tbl:ablation_GB}.
This validates the superiority of our decoding method.

\begin{figure}
	\centering
	\includegraphics[width=1\linewidth]{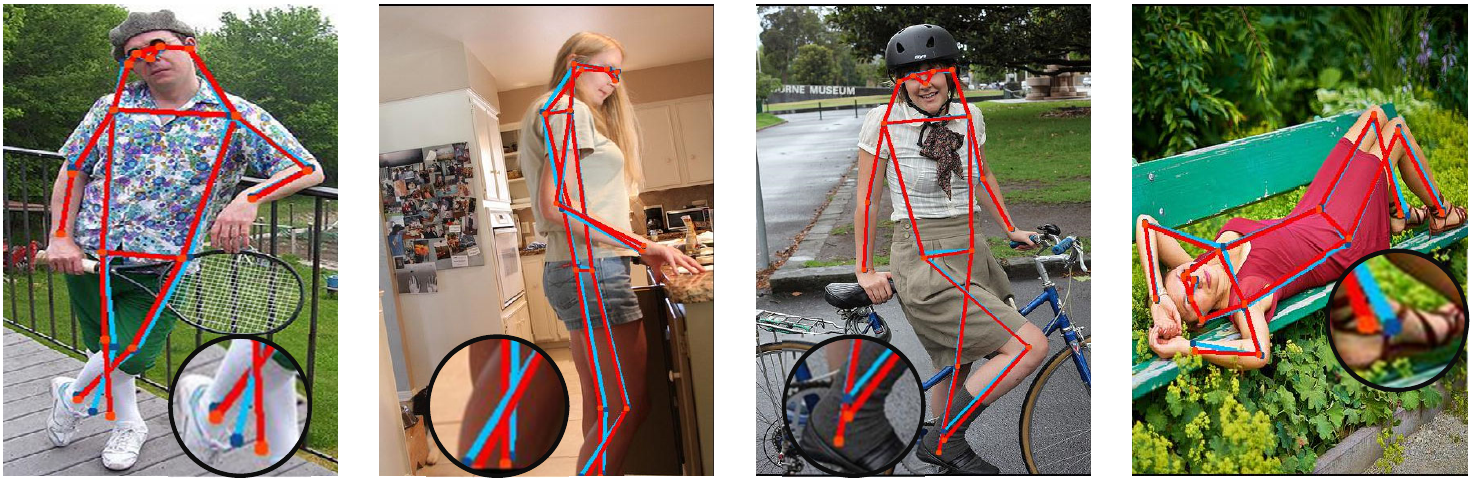}
	\caption{Qualitative evaluation
		of DARK ({\color{red}red}) {\em vs.} HRNet-W32 ({\color{cyan}cyan}) on COCO.
	}
	\label{fig:samples}
\end{figure}

\begin{table*}[htb]
	\setlength{\tabcolsep}{0.1cm}
	\begin{center}
		\caption{
			Comparison with the state-of-the-art human pose estimation methods on the COCO test-dev set. 
		}
		\label{tbl:STOA_COCO}
		\begin{tabular}{ l | c | c | c | c | c | c | c | c | c | c  }
			\hline
			Method & Backbone & Input size & \#Params & GFLOPs & $AP$ & $AP^{50}$ & $AP^{75}$ & $AP^{M}$ & $AP^{L}$ & $AR$ \\
			\hline \hline
			G-RMI & ResNet-101 & $353\times257$ & 42.6M & 57.0 
			& 64.9 & 85.5 & 71.3 & 62.3 & 70.0 & 69.7
			\\
			Integral Pose Regression & ResNet-101 & $256\times256$ & 45.1M & 11.0
			& 67.8 & 88.2 & 74.8 & 63.9 & 74.0 & -
			\\
			
			CPN & ResNet-Inception & $384\times288$ & - & -
			& 72.1 & 91.4 & 80.0 & 68.7 & 77.2 & 78.5
			\\
			RMPE & PyraNet & $320\times256$ & 28.1M & 26.7
			& 72.3 & 89.2 & 79.1 & 68.0 & 78.6 & -
			\\
			CFN & - & - & - & -
			& 72.6 & 86.1 & 69.7 & 78.3 & 64.1 & -
			\\
			CPN (ensemble) & ResNet-Inception & $384\times288$ & - & -
			& 73.0 & 91.7 & 80.9 & 69.5 & 78.1 & 79.0 
			\\
			SimpleBaseline & ResNet-152 & $384\times288$ & 68.6M & 35.6
			& 73.7 & 91.9 & 81.1 & 70.3 & 80.0 & 79.0
			\\
			HRNet & HRNet-W32 & $384\times288$ & 28.5M & 16.0
			&74.9 &\bf 92.5 & 82.8 & 71.3 & 80.9 & 80.1
			\\
			HRNet & HRNet-W48 & $384\times288$ & 63.6M & 32.9
			& 75.5 &\bf 92.5 & 83.3 & 71.9 & 81.5 & 80.5
			\\ \hline
			\bf DARK & HRNet-W32 & $128\times96$ & 28.5M & \bf 1.8
			& 70.0 & 90.9 & 78.5 & 67.4 & 75.0 & 75.9
			\\
			\bf DARK & HRNet-W48 & $384\times288$ & 63.6M & 32.9
			&\bf 76.2 &\bf 92.5 &\bf 83.6 &\bf 72.5 &\bf 82.4 &\bf 81.1
			\\ 
			\hline \hline
			G-RMI (extra data) & ResNet-101 & $353\times257$ & 42.6M & 57.0
			& 68.5 & 87.1 & 75.5 & 65.8 & 73.3 & 73.3
			\\
			HRNet (extra data) & HRNet-W48 & $384\times288$ & 63.6M & 32.9
			& 77.0 &\bf 92.7 & 84.5 & 73.4 & 83.1 & 82.0
			\\
			\hline
			\bf DARK (extra data) & HRNet-W48 & $384\times288$ & 63.6M & 32.9
			&\bf 77.4 & 92.6 &\bf 84.6 &\bf 73.6 &\bf 83.7 &\bf 82.3
			\\
			\hline
		\end{tabular}
	\end{center}
\end{table*}

\begin{table}[htb]
	\setlength{\tabcolsep}{0.1cm}
	\begin{center}
		\caption{
			Comparison on the MPII validation set.
			DARK uses HRNet-W32 (HRN32) as backbone. 
			Input size: 256$\times$256.
			Single-scale model performance is considered.
		}
		\label{tbl:SOTA_MPII_Valid}
		\resizebox{\columnwidth}{!}{%
			\begin{tabular}{ c | c | c | c | c | c | c | c | c }
				\hline
				Method & Head & Sho. & Elb. & Wri. & Hip & Kne. & Ank. & Mean \\
				\hline
				\hline
				\multicolumn{9}{c}{PCKh@0.5} \\
				\hline
				HRN32 & 97.1 & \bf 95.9 & 90.3 & 86.5 & 89.1 & \bf 87.1 & 83.3 & 90.3 \\
				\bf DARK & \bf 97.2 & \bf 95.9 & \bf 91.2 & \bf 86.7 & \bf 89.7 & 86.7 & \bf 84.0 & \bf 90.6 \\
				\hline
				\multicolumn{9}{c}{PCKh@0.1} \\
				\hline
				HRN32 & 51.1 & 42.7 & 42.0 & 41.6 & 17.9 & 29.9 & 31.0 & 37.7 \\
				\bf DARK & \bf 55.2 & \bf 47.8 & \bf 47.4 & \bf 45.2 & \bf 20.1 & \bf 33.4 & \bf 35.4 & \bf 42.0 \\
				\hline
			\end{tabular}
		}
	\end{center}
\end{table}

\vspace{0.1cm}
\noindent{\bf (ii) Coordinate encoding }
We tested how effective coordinate encoding can be.
We compared the proposed {\em unbiased} encoding 
with the standard {\em biased} encoding,
along with both the standard and our decoding method.
We observed from Table \ref{tbl:coord_encode}
that our unbiased encoding with accurate kernel centre
brings positive performance margin, regardless of 
the coordinate decoding method.
In particular, unbiased encoding contributes
consistently over 1\% AP gain in both cases.
This suggests the importance of coordinate encoding,
which again is neglected by previous investigations.

\vspace{0.1cm}
\noindent{\bf (iii) Input resolution }
We examined the impact of input image resolution/size
by testing a number of different sizes,
considering that it is an important factor relevant to model inference efficiency.
We compared our DARK model 
(HRNet-W32 as backbone) 
with the original HRNet-W32 using the 
biased heatmap supervision for training and 
the standard shifting for testing.
From Table \ref{tbl:ablation_input_size} we have a couple of observations:
{\bf (a)} With reduced input image size, as expected the model performance consistently 
degrades whilst the inference cost drops clearly.
{\bf (b)} With the support of DARK,
the model performance loss can be effectively mitigated,
especially in case of very small input resolution (\ie very fast model inference).
This facilitates the deployment of human pose estimation models
on low-resource devices, highly desired in the emerging embedded AI.

\vspace{0.1cm}
{\noindent \bf  (iv) Generality }
Besides the state-of-the-art HRNet, 
we also tested other two representative human pose estimation models
under varying CNN architectures:
SimpleBaseline \cite{xiao2018simple} 
and 
Hourglass \cite{newell2016stacked}.
The results in Table \ref{tbl:ablation_generalisation} 
show that DARK provides significant performance gain
to the existing models in most cases.
This suggests a generic usefulness
of our approach. 
We showed qualitative evaluation in Fig. \ref{fig:samples}.

\vspace{0.1cm}
{\noindent \bf  (v) Complexity }
We tested the inference efficiency impact by our method in 
HRNet-W32 at input size of $128\times 96$.
On a Titan V GPU, the running speed is reduced from 360 fps to 320 fps
in the {\em low-efficient} python environment,
\ie a drop of 11\%.
We consider this extra cost is rather affordable.

\subsection{Comparison to the State-of-the-Art Methods}

{\bf \noindent (i) Evaluation on COCO }
We compared our DARK method with top-performers
including G-RMI \cite{papandreou2017towards},
Integral Pose Regression \cite{sun2018integral},
CPN \cite{chen2018cascaded},
RMPE \cite{fang2017rmpe},
SimpleBaseline \cite{xiao2018simple},
and HRNet \cite{sun2019deep}.
Table \ref{tbl:STOA_COCO} shows the accuracy results of the state-of-the-art methods and DARK on the COCO test-dev set.
In this test, we used the person detection results from \cite{sun2019deep}.
We have the following observations:
{\bf (i)}
DARK with HRNet-W48 at the input size of 384$\times$288 achieves the best accuracy, without extra model parameters and only tiny cost increase.
Specifically, compared with the best competitor (HRNet-W48 with the same input size), 
DARK %
further improves AP by 0.7\% (76.2-75.5).
When compared to the most efficient model (Integral Pose Regression), DARK(HRNet-W32) achieves an AP gain of 2.2\% (70.0-67.8) whilst only needing 16.4\% (1.8/11.0 GFLOPs) execution cost.
These suggest the advantages and flexibility of DARK
on top of existing models
in terms of both accuracy and efficiency. 

\vspace{0.2cm}
{\bf \noindent (ii) Evaluation on MPII }
We compared DARK with %
HRNet-W32
on the MPII validation set.
The comparisons in Table \ref{tbl:SOTA_MPII_Valid} show a consistent
performance superiority of our method over the best competitor.
Under the more strict accuracy measurement PCKh@0.1,
the performance margin of DARK is even more significant.
Note, MPII provides significantly smaller training data than COCO,
suggesting that our method generalises across varying training data sizes.

\section{Conclusion}
In this work, we for the first time systematically investigated 
the largely ignored yet significant problem of 
{\em coordinate representation} (including encoding and decoding) 
for human pose estimation in unconstrained images. 
We not only revealed the genuine significance of this problem,
but also presented a novel distribution-aware coordinate representation of keypoint (DARK)
for more discriminative model training and inference.
Serving as a ready-to-use plug-in component,
existing state-of-the-art models can be seamlessly benefited from our DARK method
without any algorithmic adaptation 
at a neglectable cost.
Apart from demonstrating empirically the importance of coordinate representation,
we validated the performance advantages of DARK
by conducting extensive experiments with
a wide spectrum of contemporary models 
on two challenging datasets.
We also provided a sequence of in-depth component analysis
for giving insights on the design rationale 
of our model formulation.

{\small
	\bibliographystyle{aaai}
	\bibliography{references}
}

\end{document}